\documentclass{article}
\usepackage[preprint,nonatbib]{neurips_2020}

\usepackage{amsmath,amsthm,amssymb}
\usepackage{bm}
\usepackage{hyperref}
\usepackage{xcolor}
\hypersetup{urlbordercolor={white},}

\usepackage{booktabs}
\usepackage{multirow}
\usepackage{cleveref}

\usepackage[utf8]{inputenc} \usepackage[T1]{fontenc}    \usepackage{nicefrac}       \usepackage{microtype}      

\usepackage{pifont}\newcommand{\cmark}{\ding{51}}\newcommand{\xmark}{\ding{55}}\newcommand{\defeq}{\mathrel{\mathop:}=}

\newcommand{\rl}{\mathsf{ReLU}}

\usepackage{scalerel}
\usepackage{tikz}
\usetikzlibrary{svg.path}

\definecolor{orcidlogocol}{HTML}{A6CE39}
\tikzset{
  orcidlogo/.pic={
    \fill[orcidlogocol] svg{M256,128c0,70.7-57.3,128-128,128C57.3,256,0,198.7,0,128C0,57.3,57.3,0,128,0C198.7,0,256,57.3,256,128z};
    \fill[white] svg{M86.3,186.2H70.9V79.1h15.4v48.4V186.2z}
                 svg{M108.9,79.1h41.6c39.6,0,57,28.3,57,53.6c0,27.5-21.5,53.6-56.8,53.6h-41.8V79.1z M124.3,172.4h24.5c34.9,0,42.9-26.5,42.9-39.7c0-21.5-13.7-39.7-43.7-39.7h-23.7V172.4z}
                 svg{M88.7,56.8c0,5.5-4.5,10.1-10.1,10.1c-5.6,0-10.1-4.6-10.1-10.1c0-5.6,4.5-10.1,10.1-10.1C84.2,46.7,88.7,51.3,88.7,56.8z};
  }
}

\newcommand\orcidicon[1]{\href{https://orcid.org/#1}{\mbox{\scalerel*{
\begin{tikzpicture}[yscale=-1,transform shape]
\pic{orcidlogo};
\end{tikzpicture}
}{|}}}}

\title{Effectiveness of MPC-friendly Softmax Replacement}
\author{Marcel Keller~\orcidicon{0000-0003-2261-9376}\\ CSIRO's Data61 \\ \texttt{marcel.keller@data61.csiro.au}
\And Ke Sun~\orcidicon{0000-0001-6263-7355}\\ CSIRO's Data61 \\ \texttt{ke.sun@data61.csiro.au}}

\makeatletter\begin{document}

\maketitle

\begin{abstract}
  Softmax is widely used in deep learning
to map some representation to a probability distribution. 
  As it is based on exp/log functions that is relatively expensive in multi-party computation,
  Mohassel and Zhang (2017)
  proposed a simpler replacement based on ReLU to be used in secure computation.
  However, we could not reproduce the accuracy
  they reported for training on MNIST with three fully connected
  layers. Later works (e.g., Wagh et al., 2019 and 2021) used the
  softmax replacement not for computing the output probability distribution
  but for approximating the gradient in
  back-propagation. In this work, we analyze the two uses of the
  replacement and compare them to softmax, both in terms of
  accuracy and cost in multi-party computation. We found that the
  replacement only provides a significant speed-up for a one-layer
  network while it always reduces accuracy, sometimes significantly.
  Thus we conclude that its usefulness is limited and one should
  use the original softmax function instead.

  \emph{Changelog:} We fixed a bug in our software affecting the
  accuracy. The new figures support our conclusion more strongly.
\end{abstract}

\section{Introduction}

We use multi-class classification as a typical example where softmax is applied in deep learning.
Consider recognizing hand-written digits, where the input is
an image and the output is a class label from 0 to 9, signifying
the digit it represents. Given the input, a deep neural network
can learn a vector representation $\bm{x}=(x_0, \dots, x_9) \in \mathbb{R}^{10}$,
where a larger $x_i$ means a higher likelihood of the input being the digit
$i$. In order to turn $\bm{x}$ into a probability distribution (and to
define a learning process), the softmax function
\begin{equation}\label{eq:softmax}
p_i \defeq \frac{e^{x_i}}{\sum_j e^{x_j}}
\end{equation}
is commonly used.
It is easy to check that $\bm{p}=(p_0, p_1, \cdots)$
defines a probability distribution. Indeed, by \cref{eq:softmax}, all
entries of $\bm{p}$ are non-negative and sum up to one.

Usually, a loss function is minimized to implement learning.
For a training sample, its loss, indicating the incorrectness of the model,
is defined in terms of the output distribution $\bm{p}$
as well as the ground truth one-hot vector $\bm{y}=(y_0, y_1, \dots)$ with $y_i = 1$ for
the true class label $i$ and $y_i = 0$ otherwise.
The global loss is a sum of the per-sample losses.
A commonly used loss function is the cross-entropy:
\[ \ell \defeq -\sum_i y_i \cdot \log p_i = -\sum_i y_i \cdot \log
  \frac{e^{x_i}}{\sum_j e^{x_j}} = -\sum_i y_i \cdot x_i +
  \log \sum_j e^{x_j}, \]
which attains its minimum when $\bm{p}=\bm{y}$.
It is easy to see that a loss of zero indicates a perfect prediction.
On the other hand, assigning a small probability to the ground truth
can yield a large loss.

Finally, for the optimization process we take the partial derivative of
the loss function in every coordinate, and then perform gradient descent.
This indicates the
``correction'' on the output values needed for a better model.
In our example, this is
\begin{align*}
\bigtriangledown_i \defeq
\frac{\partial\ell}{\partial x_i} &= \frac{\partial}{\partial x_i} \Big(
                                      -\sum_k y_k \cdot x_k +
                                      \log \sum_j e^{x_j} \Big)
                                    = -y_i + \frac{e^{x_i}}{\sum_j e^{x_j}}
                                    = -(y_i-p_i).
\end{align*}
For a good model, the loss reaches a local minimum and the partial derivatives are close to
zero. Due to the derivatives of $\exp$ and $\log$, softmax appears
again in the expression of the gradient of $\ell$.
This is not generally true for any map from real vectors to
probability distributions as we will see below.

\textbf{Multi-Party Computation (MPC)} is a technology for collaborative computation without individual
parties learning the input or intermediate data. As such, it has been
proposed as a key tool for federated learning. However, the underlying
mathematics only offer modular addition and multiplication as core
operations. While it is possible to build non-linear computation using
these, the relative cost compared to the core operations is much
higher than with microprocessors. In particular for exponential
computation, there is only recent literature \cite{ACNS:AlySma19} on
how to do this compared to comparison and division \cite{FC:CatSax10},
which are the only ingredients on the softmax replacement defined
below.

\section{A Softmax Replacement}

Mohassel and Zhang \cite{SP:MohZha17} suggested to replace softmax with
\begin{equation}\label{eq:replace}
  \tilde{p}_i \defeq
  \begin{cases}
    \frac{\rl(x_i)}{\sum_i \rl(x_i)}, & \text{if~} \sum_i \rl(x_i) > 0 \\
    1/L, & \text{otherwise}
  \end{cases}
\end{equation}
where $L=\dim(\bm{x})$ is the number of possible classes and $\rl$~\cite{relu} is defined as
follows:
\[ \rl(x) \defeq \begin{cases}
    x, & \text{if~}x > 0 \\
    0. & \text{otherwise}
  \end{cases}
\]

It is easy to recognize the appeal of this function. The vector
$\tilde{\bm{p}}=(\tilde{p}_0,\tilde{p}_1,\cdots)$ in \cref{eq:replace}
is clearly a probability distribution.
Similar to \cref{eq:softmax}, it assigns the highest probability to the
largest value of $\bm{x}$.
Furthermore, piece-wise linear approximations are proven successful
in other contexts such as logistic regression. Mohassel and Zhang have
proposed to replace the sigmoid function by three-piece linear
approximation, which has shown to closely match the accuracy without the
replacement on several datasets~\cite{SP:MohZha17,hong2020privacy}.

The back-propagation implied by
using the softmax replacement as an output probability
distribution has not been spelled out in previous work.
Using the softmax replacement with cross-entropy loss
results in the following (ignoring the special case
when $\rl(x_i) = 0$):
\begin{align*}
  \tilde{\ell} \defeq -\sum_i y_i \cdot \log \tilde{p}_i &= - \sum_i y_i \cdot \log
  \frac{\rl(x_i)}{\sum_j \rl(x_j)} \\
  &= - \sum_i y_i \cdot \log \rl(x_i) + \log\Big(\sum_j
    \rl(x_j)\Big),
\end{align*}
where $y_i$ denotes the ground truth as a one-hot vector. For
back-propagation, we take the partial derivate:
\begin{align*}
\frac{\partial\tilde{\ell}}{\partial x_i} &= -y_i \cdot \frac{[x_i > 0]}{x_i} + \frac{[x_i > 0]}{\sum_j \rl(x_j)}\\
& = -[x_i > 0] \cdot \left( \frac{y_i}{x_i} - \frac{1}{\sum_j \rl(x_j)} \right)\\
& = - \frac{[x_i > 0]}{x_i} \cdot \left( y_i - \frac{\rl(x_i)}{\sum_j \rl(x_j)} \right),
\end{align*}
where $[\cdot]$ denotes the Iverson bracket (1 if the condition is true otherwise 0).
The obvious issue is division by zero and numerical instability caused by the
first term in the parentheses.
We found that
this can be fixed by defining the gradient flow
\[
\widetilde{\bigtriangledown}_i
\defeq
  \begin{cases}
    0 & y_i = 0,\ x_i < \varepsilon \\
    -1 & y_i = 1,\ x_i < \varepsilon \\
    -\left( \frac{y_i}{x_i} - \frac{1}{\sum_j \rl(x_j)} \right) & \text{otherwise}
  \end{cases}
\]
for some $\varepsilon\in(0,1)$. The reason to use a non-zero $\varepsilon$
is to limit the scale of the partial derivate. Using softmax, 
$\bigtriangledown_i$ is guaranteed to be in $(-1, 1)$, and thus we aim to constrain
$\widetilde{\bigtriangledown}_i$ in a similar range. If $x_i\ge\varepsilon$,
\begin{equation}
\left\vert \frac{y_i}{x_i} - \frac{1}{\sum_j \rl(x_j)} \right\vert
= \frac{1}{x_i}\left\vert y_i-\tilde{p}_i\right\vert
\le \frac{1}{x_i} \le \frac{1}{\varepsilon}.
\end{equation}
Therefore 
$\widetilde{\bigtriangledown}_i\in[-\frac{1}{\varepsilon},\frac{1}{\varepsilon}]$.
In our experiments, we simply fix $\varepsilon = 0.1$ and 
show that this suffices for convergence albeit with lower accuracy than softmax.
Neither $\tilde{\ell}$ nor $\widetilde{\bigtriangledown}_i$ is a continuous function
with respect to $\bm{x}$. Therefore the learning process may suffer from instability.

Kaina et al. \cite{kanai2018sigsoftmax} have suggested to mitigate the
instability using the following probability distribution:
\[
  \bar{p}_i \defeq
  \frac{\rl(x_i) + \varepsilon}{\sum_i (\rl(x_i) + \varepsilon)}
\]
for $\varepsilon = 10^{-8}$. Computing the gradient as above, we get
\begin{align*}
\frac{\partial\bar{\ell}}{\partial x_i}
& = - \frac{[x_i > 0]}{x_i + \varepsilon} \cdot \left( y_i - \frac{\rl(x_i) + \varepsilon}{\sum_j (\rl(x_j) + \varepsilon)} \right),
\end{align*}
whose absolute value is bounded by $1/(x_i + \varepsilon)$. This is 
similar to our implementation $\widetilde{\bigtriangledown}_i$.
Both have bounded gradient.

\paragraph{Replacing softmax directly in the back-propagation.}

Following Mohassel and Zhang's proposal, a number of works
\cite{PoPETS:WTBKMR21,PoPETS:WagGupCha19,NDSS:ChaRacSur20,EPRINT:PSSY20} used
the softmax replacement
directly in back-propagation. This is to say, they implement gradient
descent by manually modifying the gradient of $\ell$ with respect to $x_i$ as
\[ 
-y_i + \frac{\rl(x_i)}{\sum_i \rl(x_i)}.
\]

Taking into account the special case when $\sum_j \rl(x_j) = 0$,
we implement this approach as follows:
\[
\widehat{\bigtriangledown}_i
\defeq
  \begin{cases}
    -y_i, & \text{if~}\sum_j \rl(x_j) = 0\\
    -y_i + \frac{\rl(x_i)}{\sum_i \rl(x_i)}. & \text{otherwise}
  \end{cases}
\]
We found that it is not really necessary to treat small values of
$\sum_j \rl(x_j)$ because it is unlikely to arise for a random model.

While this is less likely to require treatment of special cases, and
it comes closer to the softmax back-propagation, the above works do
not provide a formal justification in the form of a loss
function. Nevertheless, a loss function or probability distribution is
not necessary to measure the accuracy because that can simply be done by
taking the maximum of the output values. We did so in our
implementation. However, while we managed to stabilize accuracy, it
remained considerably below either using softmax or the softmax
replacement as output probability distribution.

\section{Experiments}

We implemented training for the MNIST dataset \cite{mnist} in MP-SPDZ
\cite{CCS:Keller20} with one to three dense layers. All but the last
layer consist of 128 ReLU units~\cite{relu}. We use
fixed-point representation of fractional numbers, that is
$x \in \mathbb{R}$ is represented as $\mathsf{round}(x \cdot
2^{16})$. For rounding after multiplication, we consider two variants:
nearest and probabilistic rounding. The latter is particularly
efficient in secure computation and rounds according to proximity. For
example, 0.25 is rounded down to 0 with 0.75 probability.

To implement the exponential function we use the approach by Aly and
Smart~\cite{ACNS:AlySma19}.  They proposed to compute exponentials via
computing exponentiation with base 2 because
$e^x = 2^{x \cdot \log_2 e}$. Powers of two can be computed be
splitting the input into the integer and fractional components
$a = x + y$, where $x$ is an integer and $y \in [0,1)$. The former can be computed exactly
using bit decomposition. If $x = \sum x_i \cdot 2^i$, where $x_i=0$ or $1$, then
$2^x = \prod_i (1 - x_i + x_i \cdot 2^{2^i})$. On the other hand, $2^y$ for
$y \in [0, 1)$ can be computed using Taylor approximation. Finally,
$2^a = 2^x \cdot 2^y$.

\begin{table}
  \centering
  \caption{Time and accuracy for various models and parameters. ``$\bot$''
    stands for divergence even with the smallest possible learning
    rate. ``Rounding'' denotes the rounding after multiplication in
    fixed-point representation (probabilistic or nearest), and ``ReLU
    probability'' and ``ReLU gradient'' denote using the softmax replacement
    for output probability and the gradient, respectively.}
  \label{table:results}
  \begin{tabular}{cllrrrrrr}
    \toprule
    \multirow{2}{*}{No. layers} & \multirow{2}{*}{Rounding} & \multirow{2}{*}{Back-propagation} & \multirow{2}{*}{s/epoch} & \multicolumn{4}{c}{Accuracy after $n$ epochs} \\ \cmidrule{5-8}
    &&&& $n=5$ & $n=10$ & $n=15$ & $n=20$ \\
\midrule \multirow{6}{*}{1} & \multirow{3}{*}{Prob.} & Softmax &   12.0 & 91.6 & 92.2 & 92.2 & 92.4 \\          
&& ReLU probability &                                                     7.0 & 87.8 & 89.0 & 90.8 & 91.6 \\           
&& ReLU gradient &                                                 5.6 & 86.7 & 86.7 & 86.7 & 86.7 \\           
\cmidrule{2-8} & \multirow{3}{*}{Nearest} & Softmax &              24.3 & 91.7 & 92.1 & 92.3 & 92.5 \\          
&& ReLU probability &                                                     16.0 & 90.4 & 90.5 & 90.3 & 88.5 \\          
&& ReLU gradient &                                                 13.9 & 86.5 & 86.6 & 86.7 & 86.5 \\            
\midrule \multirow{6}{*}{2} & \multirow{3}{*}{Prob.} & Softmax &   28.2 & 95.8 & 96.9 & 97.2 & 97.6 \\          
&& ReLU probability &                                                     23.2 & 92.2 & 92.4 & 93.3 & 93.4 \\                 
&& ReLU gradient &                                                 $\bot$ & $\bot$ & $\bot$ & $\bot$ & $\bot$ \\
\cmidrule{2-8} & \multirow{3}{*}{Nearest} & Softmax &              55.3 & 96.2 & 97.2 & 97.4 & 97.5 \\          
&& ReLU probability &                                                     46.8 & 92.9 & 92.1 & 91.4 & 87.2 \\                 
&& ReLU gradient &                                                 $\bot$ & $\bot$ & $\bot$ & $\bot$ & $\bot$ \\
\midrule \multirow{6}{*}{3} & \multirow{3}{*}{Prob.} & Softmax &   33.8 & 96.7 & 97.4 & 97.7 & 97.9 \\          
&& ReLU probability &                                                     28.8 & 92.3 & 93.1 & 93.5 & 94.2 \\                 
&& ReLU gradient &                                                 $\bot$ & $\bot$ & $\bot$ & $\bot$ & $\bot$ \\
\cmidrule{2-8} & \multirow{3}{*}{Nearest} & Softmax &              70.1 & 96.8 & 97.4 & 97.5 & 97.5 \\          
&& ReLU probability &                                                     61.4 & 94.2 & 95.3 & 95.0 & 95.6 \\                 
&& ReLU gradient &                                                 $\bot$ & $\bot$ & $\bot$ & $\bot$ & $\bot$ \\
    \bottomrule
  \end{tabular}                                                                 \end{table}

Table \ref{table:results} lists our timings and accuracy results for
one run of each variant with honest-majority semi-honest three-party
computation on AWS c5.9xlarge.
Our code is available as a Docker container for
reproduction.\footnote{\url{https://github.com/mkskeller/mnist-mpc}}

Our results show that the ReLU-based softmax replacement only improves
the running time per epoch by less than 25 percent for two layers or
more while
it considerably deteriorates the accuracy for any number of
layers. If measuring the time it takes until a certain accuracy is
reached, softmax always produces the best results.
Furthermore, using the ReLU-based replacement directly in the
back-propagation does not produce convergence at all with more than one
layer.

Further notable is the fact that three layers do not improve the
accuracy for the reported number of epochs. We found that 98 percent
are achieved after 50 epochs with both two or three layers.
The same occurs in plaintext training, where
we found that a three-layer model would not improve over two
layers. This was done using the TensorFlow MNIST tutorial
\cite{mnist-tutorial} by running it as is (two dense layers) and
duplicating the first dense layer.

The three-layer model was used by
Mohassel and Zhang \cite{SP:MohZha17} and later dubbed Network A by
Wagh et al. \cite{PoPETS:WagGupCha19}. Neither makes an argument for
using three instead of two layers, however. Mohassel and Zhang claim
to reach 93.4\% accuracy after 15 epochs. They provide neither code
nor a detailed description of their protocol. We therefore lack the
information to further evaluate the considerable difference to our
accuracy results.

The recent work of Wagh et
al. \cite{PoPETS:WTBKMR21} reports a timing of 0.17 hours for 15
epochs of training, which corresponds to 41 seconds per epoch. They
improve on previous works \cite{CCS:MohRin18,PoPETS:WagGupCha19} in
the same security model, which shows that our implementation is
competitive even when using softmax.

We have also run two-party training for one dense layer with
probabilistic rounding. We found that one epoch takes 1173, 993, and
892 seconds with softmax, ReLU probality, and ReLU gradient,
respectively. However, MP-SPDZ does not support matrix multiplication
via triples generated using homomorphic encryption. This would only
benefit the dense layer computation and thus increase the relative
cost of the activation layer.

\section{Conclusion}

We conclude that the softmax replacement by Mohassel and Zhang
\cite{SP:MohZha17} is of limited use. For inference, finding the
index with the maximum value in a vector (argmax) is often enough. For
training on the other hand, the replacement proves to deteriorate the
accuracy and slow the convergence to the extent that it is more
efficient to use softmax in order to reach a certain accuracy.
Table \ref{table:conclusion} shows a comparison of the three variants
considered in this work.

\begin{table}
  \centering
  \caption{Comparison of softmax with the two replacement variants.}
  \label{table:conclusion}
  \begin{tabular}{lccccccc}
    \toprule
    \multirow{2}{*}{Variant} & \multirow{2}{*}{Gradient} & \multirow{2}{*}{Established} & \multirow{2}{*}{Differentiable} & \multirow{2}{*}{Known loss} & \multicolumn{3}{c}{Computation} \\
                             &&&&& $x<y$ & $x/y$ & $e^x$ \\
    \midrule
    Softmax  & $\bigtriangledown_i$ & \cmark & \cmark & \cmark & \cmark & \cmark & \cmark \\
    ReLU probability & $\widetilde{\bigtriangledown}_i$ & \xmark & \xmark & \cmark & \cmark & \cmark & \xmark \\
    ReLU gradient & $\widehat{\bigtriangledown}_i$ & \xmark & \xmark & \xmark & \cmark & \cmark & \xmark \\
    \bottomrule
  \end{tabular}
\end{table}

\bibliographystyle{alpha}
\newcommand{\etalchar}[1]{$^{#1}$}

\end{document}